\documentclass[lettersize,journal]{IEEEtran}

\usepackage[utf8]{inputenc} 
\usepackage[T1]{fontenc}    
\usepackage{hyperref}       
\usepackage{url}            
\usepackage{booktabs}       
\usepackage{amsfonts}       
\usepackage{nicefrac}       
\usepackage{microtype}      
\usepackage{xcolor}         
\usepackage{amsmath}
\usepackage{algorithmic}
\usepackage{algorithm}
\usepackage{makecell}

\usepackage{amsmath,amsfonts}
\usepackage{algorithmic}
\usepackage{algorithm}
\usepackage{array}
\usepackage[caption=false, font=footnotesize]{subfig}
\usepackage{textcomp}
\usepackage{stfloats}
\usepackage{url}
\usepackage{verbatim}
\usepackage{cite}
\usepackage{graphicx}
\usepackage{epstopdf}
\usepackage[utf8]{inputenc}
\usepackage[T1]{fontenc}
\usepackage{booktabs} 
\usepackage{diagbox}  
\usepackage{multirow} 
\usepackage{threeparttable} 
\usepackage{epstopdf} 
\usepackage{colortbl}  
\definecolor{codegreen}{rgb}{0,0.5,0}
\definecolor{codeblue}{rgb}{0,0,0.9}
\definecolor{codeblues}{rgb}{0,0,0.4}
\definecolor{codegray2}{rgb}{0.4,0.4,0.4}
\definecolor{codegray}{rgb}{0.9,0.9,0.9}
\usepackage{fontawesome}
\begin{document}
\title{Spiking Neural Network for Ultra-low-latency and High-accurate Object Detection}
\author{Jinye Qu, Zeyu Gao, Tielin Zhang, Yanfeng Lu,~\IEEEmembership{Member,~IEEE},  Huajin Tang,~\IEEEmembership{Member,~IEEE}, \\and Hong Qiao,~\IEEEmembership{Fellow,~IEEE}
\thanks{This work is supported by National Key Research and Development Plan of China (Grant 2020AAA0105900), and partially supported by Beijing Natural Science Foundation (Grant L211023) and National Natural Science Foundation of China (Grants 91948303, 61627808). (Jinye Qu and Zeyu Gao are co-first authors.) \textit{(Corresponding author: Yanfeng Lu.)}
\quad

Jinye Qu, Zeyu Gao, Yanfeng Lu, and Hong Qiao are with the State Key Laboratory of Multimodal Artificial Intelligence Systems, Institute of Automation, Chinese Academy of Science (CASIA), Beijing 100190, China, and also with the University of Chinese Academy of Sciences (UCAS), Beijing 100049, China (e-mail: yanfeng.lv@ia.ac.cn).

Tielin Zhang is with the Laboratory of Cognition and Decision Intelligence for Complex Systems, Institute of Automation, Chinese Academy of Sciences (CASIA), Beijing 100190, China, and also with the School of Artificial Intelligence, University of Chinese Academy of Sciences (UCAS), Beijing 100049, China (e-mail: tielin.zhang@ia.ac.cn).  

Huajin Tang is with the College of Computer Science and Technology, Zhejiang University, Hangzhou 310027, China (e-mail: htang@zju.edu.cn).
\quad 
}
}

\maketitle

\begin{abstract}
Spiking Neural Networks (SNNs) have garnered widespread interest for their energy efficiency and brain-inspired event-driven properties. While recent methods like Spiking-YOLO have expanded the SNNs to more challenging object detection tasks, they often suffer from high latency and low detection accuracy, making them difficult to deploy on latency sensitive mobile platforms. Furthermore, the conversion method from Artificial Neural Networks (ANNs) to SNNs is hard to maintain the complete structure of the ANNs, resulting in poor feature representation and high conversion errors. To address these challenges, we propose two methods: timesteps compression and spike-time-dependent integrated (STDI) coding. The former reduces the timesteps required in ANN-SNN conversion by compressing information, while the latter sets a time-varying threshold to expand the information holding capacity. We also present a SNN-based ultra-low latency and high accurate object detection model (SUHD) that achieves state-of-the-art performance on nontrivial datasets like PASCAL VOC and MS COCO, with about remarkable 750× fewer timesteps and 30\% mean average precision (mAP) improvement, compared to the Spiking-YOLO on MS COCO datasets. To the best of our knowledge, SUHD is the deepest spike-based object detection model to date that achieves ultra low timesteps to complete the lossless conversion.
\end{abstract}



\section{Introduction}
\IEEEPARstart{W}{ith} the development of high-performance computing devices, artificial neural networks (ANNs) have made achievements in many artificial intelligence tasks such as image classification~\cite{krizhevsky2017imagenet}, object detection~\cite{redmon2016you}, and sequential decision-making~\cite{silver2016mastering} in recent years. However, ANNs has a huge energy consumption, which makes it difficult to deploy on mobile devices.
Spiking Neural Networks (SNNs) is the third-generation artificial neural network~\cite{maass1997networks}. Inspired by biological neurons~\cite{mainen1995reliability}, it changes the complex multiplication operation in the ANNs into a simple accumulation operation and transmits information through spikes sequences~\cite{9747018}\cite{9746792}. Due to the sparsity of spike events and the characteristics of event-driven computing, SNNs have remarkable energy efficiency and are the neural network of choice for neuromorphic chips~\cite{merolla2014million,indiveri2011neuromorphic,basu2018lowpower,kim2018emerging}. It is generally believed that SNNs has greater development potential and bionic value. 

Because of the non-differentiation of SNNs, the gradients cannot be computed directly during backward propagation, which makes it difficult to obtain SNNs by training directly. There are currently two main methods of obtaining SNNs, one is obtained by conversion~\cite{spiking-yolo,initialmemvoltage,yuqiang2022constructing}, i.e. processing the trained ANNs weights to obtain available SNNs, and the other one is learning SNNs weights from scratch~\cite{yao2022glif}~\cite{zhangtielin2022tunning} by spike-time-dependent plasticity (STDP)~\cite{diehl2015unsupervised,lew2020early,xiang2019STDP} or spike-time-dependent backpropagation (STDB)~\cite{rathienabling, shrestha2018slayer}. On that basis, a variety of correlation optimization algorithms are derived~\cite{ijcai2022p343,AxonalDelay,ijcai2022p347}. The approach to learning from scratch requires a significant amount of time and computer resources relative to the conversion approach. The conversion approach takes full advantage of the ease of training of ANNs and can promptly obtain usable weights from the trained ANNs~\cite{9746566}. Both methods are widely used in shallow SNNs with good results. Q. Yu et al. used a double threshold scheme and an augmented spike scheme to achieve lossless conversion in MNIST, FashionMNIST and CIFAR10~\cite{yuqiang2022constructing}. C. Hong et al. proposed a modified SpikeProp learning algorithm, which
ensures better learning stability for SNNs\cite{hong2020training}. N. Rathi et al. proposed using the first convolutional layer as the coding layer and using a gradient descent-based training method to make the SNN accuracy close to that of an ANN with the same structure~\cite{rathi2021dietsnn}. At the same time, many works have been devoted to the use of SNNs on deeper networks. J. Ding et al. achieved an ANN to SNN conversion with a loss of 0.8\% using a PreActResNet-34 network on the CIFAR-100 dataset~\cite{huang2021optimal}. Y. Li et al. achieved a 0.23\% accuracy loss ANN to SNN conversion on the MS COCO dataset using resnet50~\cite{li2022spike}. Y. Hu et al. attempted the conversion on a deep resnet network and achieved an accuracy loss of about 1.16\% under 50 layers on ImageNet dataset~\cite{hu2021spiking}. These works pushed the development of SNNs to deeper networks.

In the past, SNNs were mainly applied to simple tasks such as image classification~\cite{initialmemvoltage,liu2022spikeconverter,yuqiang2022constructing,zhangmalu2022rectified}. In recent years, some works have used tried to promote SNNs to more challenging tasks, such as multi-sensory integration learning~\cite{zhangtielin2022motif}, object detection~\cite{spiking-yolo}, reinforcement learning~\cite{zhang2022multi}. One of the most concerned directions is the SNN based object detection.
Spiking-YOLO~\cite{spiking-yolo} is the first SNN based object detection model that pushes the boundaries of the field by achieving a near lossless ANN to SNN conversion at timesteps = 8000 based on a YOLOv3-tiny backbone network. Nevertheless, the practicality and deployment of this network is hampered by its excessive timesteps requirement and the depth of 23 layers limits its detection performance.
FSHNN~\cite{chakraborty2021fully} combines the STDP, STBP, and Monte Carlo Dropout methods and makes FSHNN exceed the accuracy of its heterogeneous ANN based RetinaNet. It has reduced timesteps from thousands to 300, significantly improving energy efficiency. However, 300 timesteps is still not sufficient for the model to run on mobile robot platforms and the model's object detection accuracy on the COCO dataset still does not match the performance of current mainstream ANN models, such as YOLOv5.

In summary, most of the previous works~\cite{spiking-yolo,rueckauer2017conversion,patel2021spiking,initialmemvoltage,Spiking-maxpooling,ijcai2022p347} require tremendous timesteps to reach lossless conversion, which makes it difficult to deploy SNNs on the latency-sensitive mobile terminal devices. The excessively slow processing speed is also hardly acceptable for real-time object detection. In addition, the current ANN-SNN conversion methods do not apply to all ANN structures, which may cause some structural damage of the ANN partly during the conversions, reducing the accuracy of the ANN, which equates to a reduction in the accuracy of the converted SNN. So to convert the complete structure of a deep neural network into a SNN with both ultra-low timesteps and high accuracy is a significant issue.

To overcome the challenges mentioned above, we introduce two novel methods to reduce time latency while maintaining comparable accuracy and low energy cost: timesteps compression and spike-time-dependent integrated coding. Further, we present an low-latency and high-accurate SNN based object detection model called SUHD. 
Our contributions can be summarized as follows:

	\begin{itemize} 
    \item{A timesteps compression method is proposed, which compresses multi-timesteps into one timestep, reducing timesteps requirements for ANN to SNN conversion and inference, providing the possibility of SNNs deployment for engineering applications. Compared to Spiking-YOLO, we are able to reduce the timesteps requirement by more than 750 times with comparable accuracy.}
	\end{itemize} 
	
	\begin{itemize} 
	\item{We propose a spike-time-dependent integrated coding (STDI) method and implement it using a time-varying threshold neuron model. This approach further reduces the inference time of SNN by approximately 38\%, which is mainly caused by the increased information capacity of individual spikes.}
	\end{itemize} 

	\begin{itemize}
          \item{The proposed Spike-SPPF structure and Spike-Maxpooling method provide a way to address the conversion issue of Spatial Pyramid Pooling - Fast (SPPF) structures, achieving lossless conversion of the Maxpool layer with any stride and realizing a lossless conversion of SPPF.}
	\end{itemize} 
    

    \begin{itemize}
    \item{Based on the methods mentioned above, we propose a ultra-low-latency and high-accurate SNN based object detection model, called SUHD. SUHD has demonstrated excellent performance on two challgening datasets (PASCAL VOC and MS COCO), achieving state-of-the-art results with 4 timesteps.}
    \end{itemize}

\section{Related Works}

\subsection{Neuronal Coding}
Frequency coding is a widely used coding method that relies on spike firing rates within a certain timesteps to convey information. However, due to the binary nature of the spikes, the spike firing ratio does not transmit information very efficiently. As a result, when encountering complex information, frequency coding must ensure accurate transmission of the information at the cost of large timesteps. 

Temporal coding is an advanced coding method that embeds time information into the spike train. The combination of time information and frequency information allows the spike train to carry more information. It includes time-to-first-spike~\cite{thorpe2001spike,oh2022neuron}, rank-order coding ~\cite{thorpe1990spike}, and phase coding~\cite{phase-coding}. It has achieved remarkable results in deep SNNs.
\subsection{Conversion Methods}
The ANN to SNN conversion has been one of the hottest issues in the last few years. Many methods have been proposed and some significant developments have been achieved based on the conversion methods. The subtraction reset~\cite{rueckauer2017conversion} discards the fixed reset potential and retains the spike intensity information.
The temporal-separation (TS)~\cite{liu2022spikeconverter}
 are proposed in 2022, which eliminates errors caused by incorrect firing order of negative spikes. TS separated the accumulation and firing phases, achieving lossless conversion in simple models. T. Bu et al. proposed the theory of membrane potential initialization~\cite{initialmemvoltage} and pointed out that the membrane potential can be initialized to half of the threshold value, which can achieve lossless conversion in sufficiently large timesteps. 
The conversion of deep ResNet~\cite{hu2021spiking} was proposed by Y. Hu et al. to solve the conversion problem of bottleneck structures.
\subsection{Spiking-YOLO}
Spiking-YOLO~\cite{spiking-yolo} is the first attempt to convert ANN to SNN to achieve object detection, which could get relatively high accuracy based on very large timesteps. It uses channel-wise normalization and signed neurons featuring imbalanced threshold to reduce the conversion error and is based on YOLOv3-tiny for conversion. At timesteps = 3000 on MS COCO dataset, it achieves close to lossless performance with an mAP of 25\%. However, we found during implementation that the large timesteps required significant computational resources, making it impossible to use on common PC or mobile robotics platform. Additionally, the actual computing speed is extremely slow.

\section{Methods}
In this section, we present our systematic approach to converting ANNs to SNNs with competitive accuracy and reduced timesteps. In subsection A, we conduct a detailed analysis of the mathematical processes involved in ANN to SNN conversion, identifying the main sources of errors. In subsections B and C, we introduce our novel methods of timesteps compression and spike-time-dependent integrated coding, which help eliminate these errors. Finally, in subsection C, we implement the conversion of the SPPF structure and apply our approaches to construct a state-of-the-art SUHD object detection model.
\label{sec:methods}
\subsection{ANN to SNN Conversion Algorithm and Error Analysis}
\label{sec:ANN to SNN conversion analysis and error}

First of all, we implemented an ANN neuron model base for a single neuron as follows:
\begin{equation}
\label{eq:ANN equation}
    {x^l} = \sum\limits_{i = 1}^n {w_i^{l - 1}x_i^{l - 1} + b_i^l},
\end{equation}

where $x^l$ denotes the output value of the neuron in layer $l$, $w_i^{l-1}$ denotes $i$-th weight of layer $l-1$ to layer $l$, $x_i^{l-1}$ denotes the output value of $i$-th neuron in layer $l-1$, and $b_i^l$ denotes the bias of the output neuron.

The LIF and IF models are two basic biological neuronal models. The former has a continuous leaky current in the absence of stimulation, allowing the membrane potential to gradually return to resting potential, which may lead to information loss. In order to accurately represent the recurrent relationships between neurons, we adopted the IF neuron model as the basic neuron model.

Meanwhile, we use the spiking ratio for the SNN neuron model to substitute the simulated value $x$ in Eq. \ref{eq:ANN equation}. To establish the corresponding equivalence, we begin with the input potential of the SNN neuron and analyze its dynamics.
\begin{equation}
    z^l(t) =  {\sum\limits_{i = 1}^n {w_i^{l - 1}s_i^{l - 1}(t) +  {b_i^l} } }.
    \label{eq:3}
\end{equation}

where $z^l(t)$ represents input of that neuron in layer $l$ at time $t$, $s_i^{l - 1}(t) $ represents the spike of $i$-th neuron of layer $l-1$, it can be formulated as follows:
\begin{equation}
    \label{eq:s}
    s=\left\{
    \begin{aligned}
    1 & , & V_{mem} \geq V_{thr}, \\
    0 & , & V_{mem} \textless V_{thr}, 
    \end{aligned}
    \centering
\right.
\end{equation}

the threshold $V_{thr}$ was set to 1. 
Then we can denote the sum of the input voltage of that neuron as follows:
\begin{equation}
    U_{in}^l = \sum\limits_{t = 1}^T {\sum\limits_{i = 1}^n {w_i^{l - 1}s_i^{l - 1}(t) + \sum\limits_{t = 1}^T {b_i^l} } },
    \label{eq:Uin}
\end{equation}

then we denote the spiking ratio as ${r^l}$ and denote output potential of that neuron in layer $l$ as ${U_{out}^l}$, Thus its numerical relationship can be represent as:
\begin{equation}\label{eq:4}
    r_{}^l = \frac{{\sum\limits_{t = 1}^T {s_{}^l(t)} }}{T} = \frac{{U_{out}^l}}{T},
\end{equation}

to limit the spiking ratio to a reasonable [0, 1], we performed channel-wise weights and bias normalization operations~\cite{spiking-yolo} to scale the activation values:
\begin{equation}\label{eq:6}
{w'}_{i}^{l} = \frac{{w_i^l \times \max_{in}}}{{\max_{out}}},\\
{b'}_{i}^{l} = \frac{{b_i^l \times \max_{in}}}{{\max_{out}}},
\end{equation}

where max$_{in}$ and max$_{out}$ denote the maximum input and output activation values of the current channel. 
Ideally, the voltage input to the neuron and the voltage output from the neuron should be the same, i.e. ${U_{in}} = {U_{out}}$~\cite{liu2022spikeconverter}. Thus combining the Eq. \ref{eq:Uin} and Eq. \ref{eq:4}, we will get:
\begin{equation}\label{eq:7}
{r^l} = \sum\limits_{i = 1}^n {{w'}_i^{l - 1}r_i^{l - 1} + {b'}_i^l},
\end{equation}

it can be seen from Eq. \ref{eq:7} that in an ideal state, the SNN replaces the analog value in the ANN with spiking ratio for lossless conversion and information transfer.
For convenience, we use $w$ and $b$ to denote the normalized weights and biases, as shown in Eq. \ref{eq:6}.
Thus, we can obtain the relationship between the membrane potential at time $T$ and $0$:
\begin{equation}
\begin{aligned}
    V_{mem,i}^l(T) = V_{mem,i}^l(0) + \sum\limits_{t = 1}^T {\sum\limits_{i = 1}^n {w_i^{l - 1}s_i^{l - 1}(t)} } + \\
   \sum\limits_{t = 1}^T {b_i^l}-\sum\limits_{t=1}^T{s^l_i(t)},
\end{aligned}
\label{eq:9}
\end{equation}

where $V_{mem,i}^l(t)$ represent the membrane potential of the $i$-th neuron of layer $l$ in time $t$.

\subsubsection{Tremendous timesteps demand}
\label{sec:Quantization error}
\begin{figure}[htbp]
    \centering
    \includegraphics[width=0.5\textwidth]{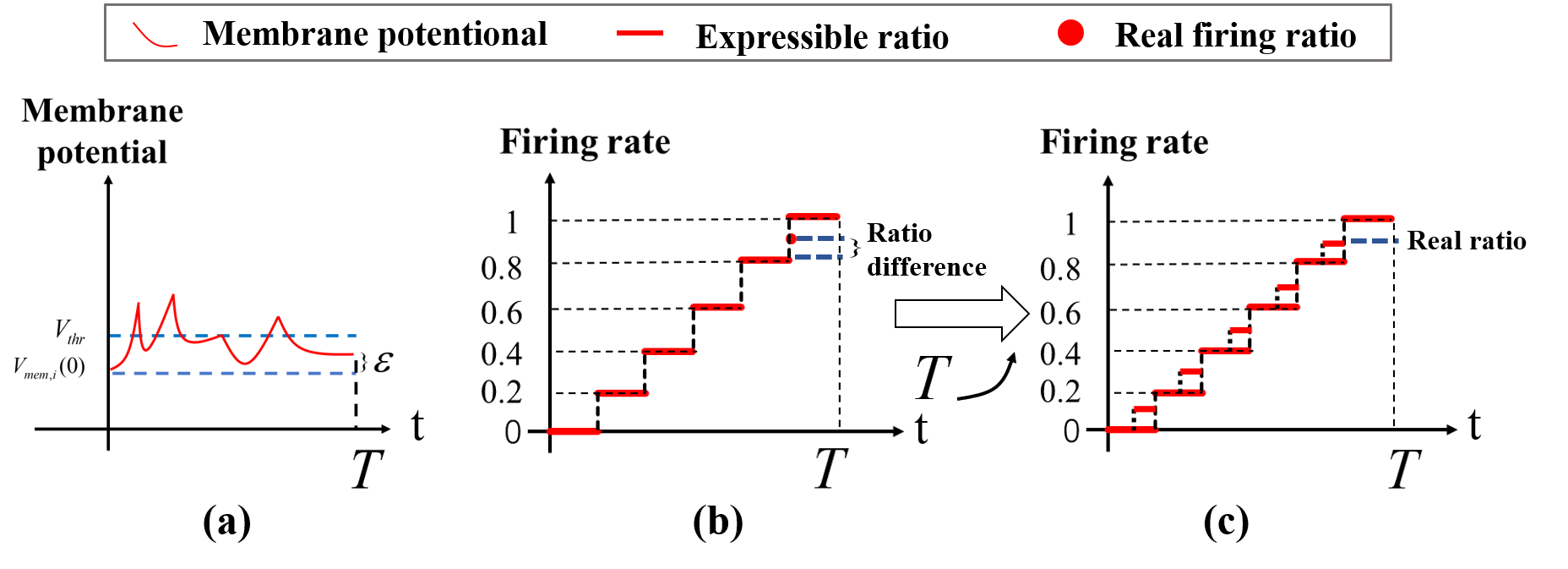}
    \caption{Quantization errors. Assuming timesteps ($T$) is 5. In (a), the membrane potential does not reach the spiking threshold at the final $t=T$ moment, leading to the potential remaining in the cell membrane without being transmitted. The equivalence to (b), only 6 values can be represented in this process: 0, 0.2, 0.4, 0.6, 0.8, 1.0. Therefore, many values cannot be expressed precisely. In (c), we improve the density of expressed values by increasing $T$, resulting in a more accurate expression of the spiking ratio. }
    \label{fig:Residual potential}
\end{figure}

Combining Eq. \ref{eq:4} and Eq. \ref{eq:9}, Then we can get:
\begin{equation}\label{eq:14}
r_i^l = \frac{(V_{mem,i}^l(0) - V_{mem,i}^l(T))}{T} + \sum\limits_{i = 1}^n {{w_i}r_i^{l - 1}} + b_i^l,
\end{equation}

ideally, in order to satisfy $U_{in}=U_{out}$ and the equivalence of Eq. \ref{eq:7}, $V_{mem,i}^l(0) - V_{mem,i}^l(T)$ should be 0. But, from Eq. \ref{eq:9},
we can see that $\sum\limits_{t = 1}^{T} s_i^l(t) $ is a step function. For $V_{mem,i}^l(0) - V_{mem,i}^l(T) = 0$
 to be true, the $\sum\limits_{t = 1}^T {\sum\limits_{i = 1}^n {w_i^{l - 1}s_i^{l - 1}(t)} } + \sum\limits_{t = 1}^T {b_i^l}$ has to be the same step function. It certainly doesn't always satisfy this condition. Therefore:

\begin{equation}
V_{mem,i}^l(0) = V_{mem,i}^l(T) - \varepsilon, (\varepsilon < {V_{thr}}),
\end{equation}

the $\varepsilon $ is the residual membrane potential and the $\frac{\varepsilon}{T}$ is called quantization error, as shown in Fig. \ref{fig:Residual potential}. Referring to the Eq. \ref{eq:14} and Fig. \ref{fig:Residual potential}(b), previous works have mainly focused on scaling down the quantization error by increasing T~\cite{spiking-yolo,burst,chakraborty2021fully}, as shown in Fig. \ref{fig:Residual potential}(c).
Since most of the current SNN-based works focus on image classification tasks, this approach is feasible. However, the object detection task requires more accurate position regression, which requires the spike sequence to have a very accurate numerical representation. Besides, most of such tasks are based on deep SNNs, which exacerbates quantization errors. Thus, we should often greatly improve the timesteps $T$ to cope with the high accuracy of the numerical expression, which takes huge timesteps for conversion (256-8000)~\cite{spiking-yolo,chakraborty2021fully,burst}.
However, the timesteps are not infinitely long, because it brings a huge amount of energy consumption and a huge running time demand. This defeats the original purpose of real-time object detection. We believe that delivering more information in limited timesteps is an important direction for SNNs development.\\



\subsubsection{Truncation error}
\label{sec:Truncation error}
\begin{figure}[htbp]
    \centering
    \includegraphics[width=0.45\textwidth]{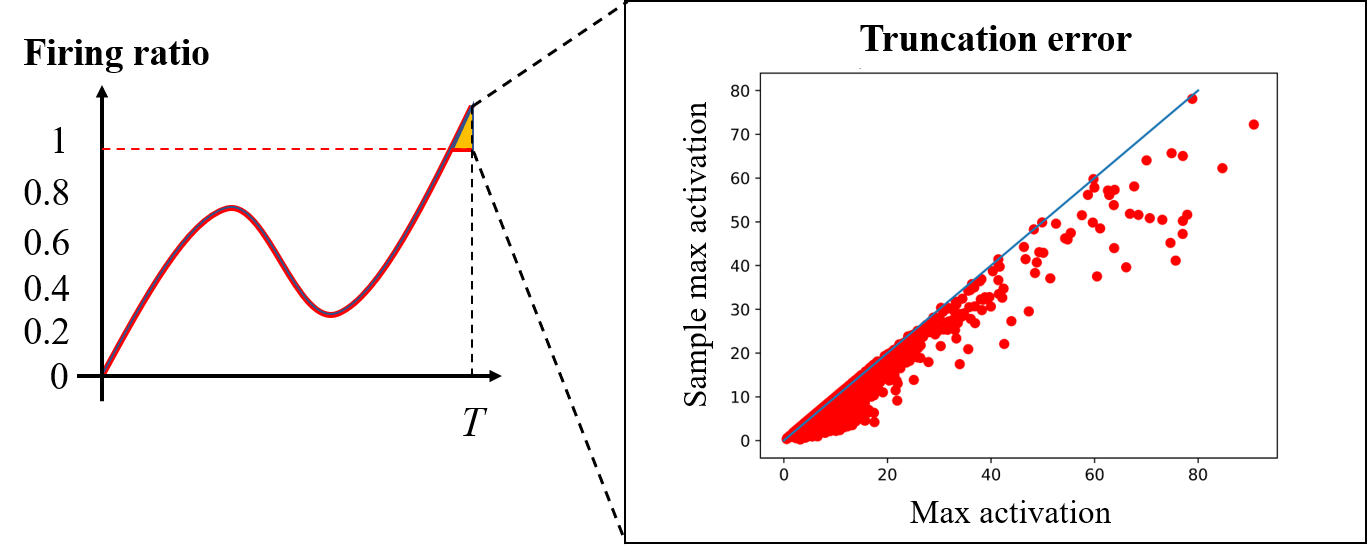}
    \caption{Truncation errors. The image on the left illustrates the generation of truncation errors, where messages exceeding the upper limit of the firing ratio are truncated by default. In the image on the right, we randomly selected 100 images from the COCO dataset, measured their maximum activation value at each layer in the model, and represented that activation value on the horizontal axis. At the same time, we measured the maximum activation value at each layer using the ANN to SNN conversion sample dataset, and represent this activation value on the vertical axis. The blue slash indicates the ideal situation when the actual maximum activation is equal to the maximum activation of the sample data. The red dots are the relative positions of the activation values for the layer under different samples.}
    \label{fig:Truncation errors}
\end{figure}

In Eq. \ref{eq:6}, to reduce the spiking ratio to a reasonable range, we scale the weight based on the maximum activation value\cite{spiking-yolo}. The maximum activation value is taken from the samples which is often based on a part of the conversion dataset. In most cases, this maximum activation value is applicable. However, in actual detection cases, the activation value may still be greater than the maximum value in the sample, as shown in Fig. \ref{fig:Truncation errors}. Therefore, for real applications in object detection tasks, $r^l$ may be greater than the spiking ratio upper limit. At this time, the part higher than 1 is cut out by default. This error reduces SNN’s accuracy.
\subsection{Timesteps Compression}
\label{sec:timesteps compression}
The large timesteps can reduce the quantization error, unevenness error, and thus the conversion loss. However, larger timesteps also bring an increase in time latency.
We propose the timesteps compression method to alleviate this problem. Timesteps compression means compressing information from multiple timesteps into one timestep and delivering the information using burst or binary spikes. Burst spikes\cite{connors1990intrinsic,izhikevich2003bursts,lisman1997bursts} are a set of short inter-spike interval (ISI) spikes that can issue multiple spikes at one timestep. 
\begin{figure*}[htbp]
    \centering
    \includegraphics[scale=0.7]{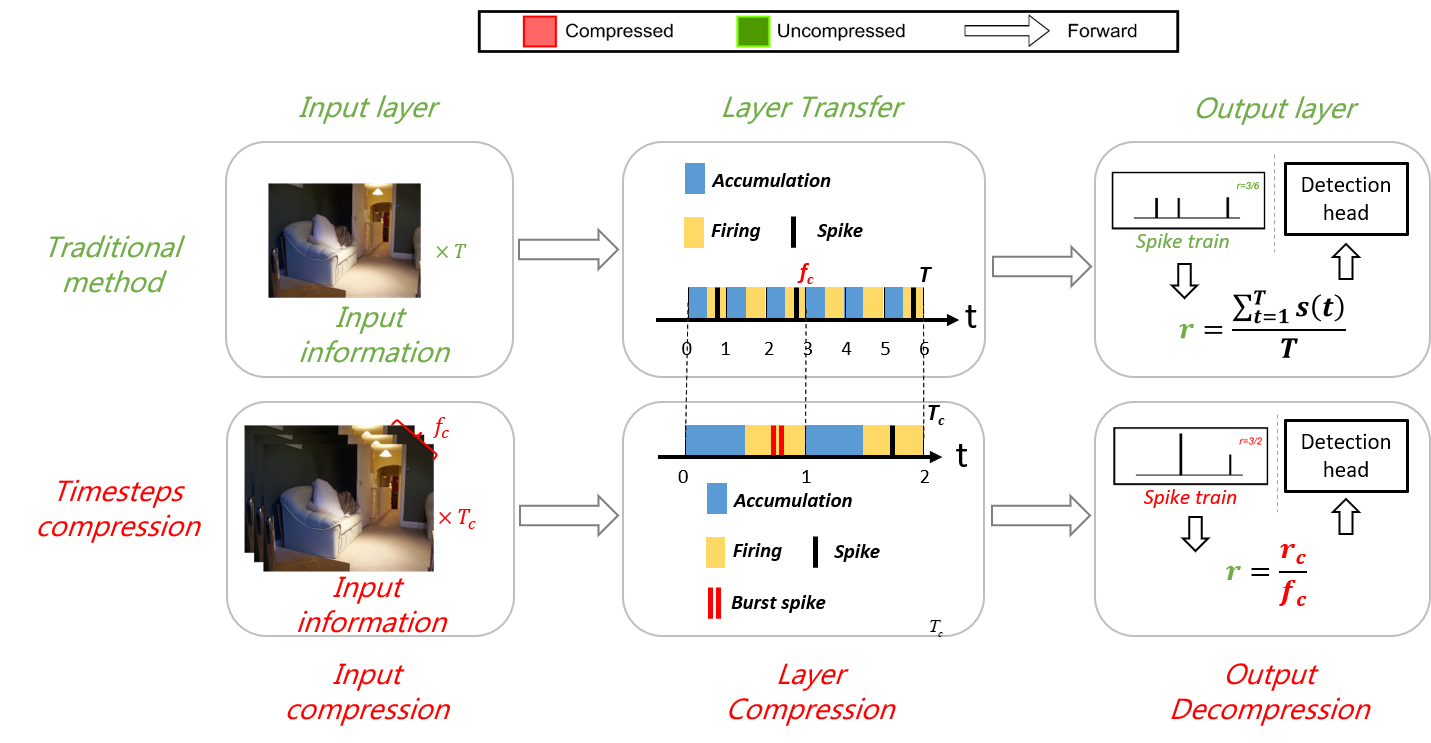}
    \caption{Process comparison between proposed timesteps compression and traditional methods. We first denote the compression scale as $f_c$, which indicates that each compressed timestep contains information from $f_c$ uncompressed timesteps. For the sake of an example, the $f_c$ is set to 3. And then we denote compressed timesteps and firing ratio as $T_{c}$ and $r_{c}$.}
    \label{fig:timesteps compression}
\end{figure*}

As shown in Fig. \ref{fig:timesteps compression}, we represent the fully equivalent uncompressed and compressed processes. timesteps compression consists of i) input compression, ii) layer compression, and iii) output decompression. We assume that the timesteps have a compression scale of $f_c$. 
Input compression compresses information from $f_c$ timesteps into one timestep and reduces the timesteps to $\frac{1}{f_c}$. This operation makes the input information the same as that with the uncompressed state. In layer compression, the compressed and uncompressed inputs within a single timestep are related as follows:
\begin{equation}
    \label{IFMODELaccumulation}
    z_c=\sum_{t=1}^{f_c}z(t)
    =\sum_{i=1}^n\sum_{t=1}^{f_c}w_i^{l-1}s_i^{l-1}(t)+f_cb,
\end{equation}

where $z_c$ represents the compressed input, $s_i^{l-1}$ denotes the output spikes of $i$-th neuron in previous layer. Multiple inputs lead to changes in the spike firing as follows:
\begin{equation}
\centering
    \label{IFMODELfiring}
    s_c=\left\{
    \begin{aligned}
    min(k,f_c) & , & V_{mem} \geq k*V_{thr}, \\
    0 & , & V_{mem} \textless V_{thr},
    \end{aligned}
    \centering
\right.
\end{equation}

 $s_c$ represents the spike issued, and when the $k \textgreater 1$, the $s_c$ is burst spike.
Layer compression allows multiple spikes to be issued in one timestep, thereby increasing the density of information within a timestep.
Ideal mathematical relationships can be presented as follows:

\begin{equation}
    U_{in}^l = \sum\limits_{i = 1}^n{\sum\limits_{t = 1}^{T_c}  {w_i^{l - 1}s_{c,i}^{l - 1}(t) + {{T_c}{f_c}b_i^l}} },
    \label{eq:Uinofcompression}
\end{equation}
\begin{equation}
    r_c^l = \frac{{U_{in}^l}}{T_c} = \sum\limits_{i = 1}^n {w_i^{l - 1}\frac{{\sum\limits_{t = 1}^{T_c}{s_{c,i}^{l - 1}(t)}}}{{T_c}}} + {f_c}b_i^l,
\end{equation}

input compression and layer compression are equivalent to compressing multiple timesteps into one timestep, which is the core of the timesteps compression. 
This relationship is illustrated in the layer compression stage and layer transfer stage in Fig. \ref{fig:timesteps compression}.
By input compression and layer compression, the firing ratio is mapped from $r\in[0,1]$ to $r_c\in[0,f_c]$, and the oversized firing ratio does not match the firing ratio at $f_c$ times uncompressed timesteps. Output decompression solves this problem by remapping the firing ratio from $r_c\in[0,f_c]$ to $r\in[0,1]$ as shown in Eq. \ref{eq:11}.




\begin{equation}
\label{eq:11}
\begin{aligned}
r^l = \frac{r_c^l}{f_c} = \sum\limits_{i = 1}^n {w_i^{l - 1}\frac{{\sum\limits_{t = 1}^{T_c}{s_{c,i}^{l - 1}(t)}}}{{T}}} +  b_i^l \\
= \sum\limits_{i = 1}^n {w_i^{l - 1}\frac{{\sum\limits_{t = 1}^{T}{s_{i}^{l - 1}(t)}}}{{T}}} +  b_i^l,
\end{aligned}
\end{equation}

as shown above, timesteps compression improved the information-carrying capacity of a single timestep and reduce the quantization error. The experiments in Sec. \ref{sec:Optimization of Accuracy with Different Methods} confirm this inference.\\

\subsection{Spike-Time-Dependent Integrated Coding}
\label{sec:Pulse intensity group counting method}
Recent object detection researches have increasingly emphasized real-time and energy efficiency. Although in the previous section, we used timesteps compression to reduce the requirement for timesteps in the model inference process, the use of frequency coding makes the model still inefficient. At the same time, it has been reported that neurons coordinate action potentials in different ways even when the spike firing rates are the same. These reports suggest that intercellular communication is the result of a combination of various coding methods. 

In particular, we use TS~\cite{liu2022spikeconverter} in our work to avoid the damage to accuracy caused by the negative spikes of misordered firing. The use of TS concentrates the spikes at the beginning of the timesteps, which gives us the opportunity to embed temporal information in the spike train.
Therefore, we propose an encoding method called the spike-time-dependent integrated (STDI) coding method, which further improves the inference speed and energy efficiency of the model.\\
\subsubsection{Weighted Spikes}
We first attach weights to the spikes based on their firing time. The weights are defined as follows:
\begin{equation}
    \tau(t) = T-t+1,
    \label{eq:spikeweight}
\end{equation}

where $t$ denotes the time of spike firing over whole timesteps. After the weights are defined, the input value represented by a firing spike becomes $s(t)*\tau$. Fig. \ref{fig:STDIcoding}(a) shows the correspondence between the weighted spikes and the input values. Fig. \ref{fig:STDIcoding}(a) also expresses that STDI can make the spike firing ratio within $t\in [0, T]$ much greater than 1, which resolves the truncation error. When faced with excessive truncation errors, STDI can eliminate such errors by integrating multiple spikes (burst spikes) in one timestep, as shown in Fig. \ref{fig:STDIcoding}(a) when the input value is 12.
Weighted spikes bring about a change in the way input information is encoded. In previous SNN object detection works, we needed to keep constant value information input throughout the timesteps and then encode it using the first layer of the SNN model. With spikes being weighted, the information only needs to be fed into the model at the first timestep. This reduces energy consumption to a certain extent.

\subsubsection{Model Adaption for STDI}
Due to the weighted spikes, we need to make appropriate adjustments to the IF model. 
Firstly, the threshold $V_{thr}$ was adjusted for weighted spikes as Eq. \ref{eq:variablethreshold}.
\begin{equation}
    \label{eq:variablethreshold}
    V_{thr}=\tau(t)*v_{thr},
\end{equation}

the $v_{thr}$ denotes the initial threshold, often set to 1.
Then we adjust the input as follows:
\begin{equation}
    z^l =  {\sum\limits_{i = 1}^n {w_i^{l - 1}s_i^{l - 1}(t)*\tau (t)+  {b_i^l} } },
    \label{eq:STDIinput}
\end{equation}

therefore Eq. \ref{eq:Uin} and Eq. \ref{eq:4} must be changed to:
\begin{equation}
    \label{eq:uin of spike-time-dependent integrated coding method}
    U_{in}^l = \sum\limits_{t = 1}^T {\sum\limits_{i = 1}^n {w_i^{l - 1}s_i^{l - 1}(t)*\tau(t) + \sum\limits_{t = 1}^T {b_i^l} } },
\end{equation}
\begin{equation}
    \label{eq:fasttimedecoder}
    r^l= \sum\limits_{i = 1}^n {w_i^{l - 1}\frac{{\sum\limits_{t = 1}^{T}{s_{i}^{l - 1}(t)*\tau(t)}}}{{T}}} +  b_i^l,
\end{equation}

\begin{figure*}[htbp]
    \centering
    \includegraphics[width=1.0\textwidth]{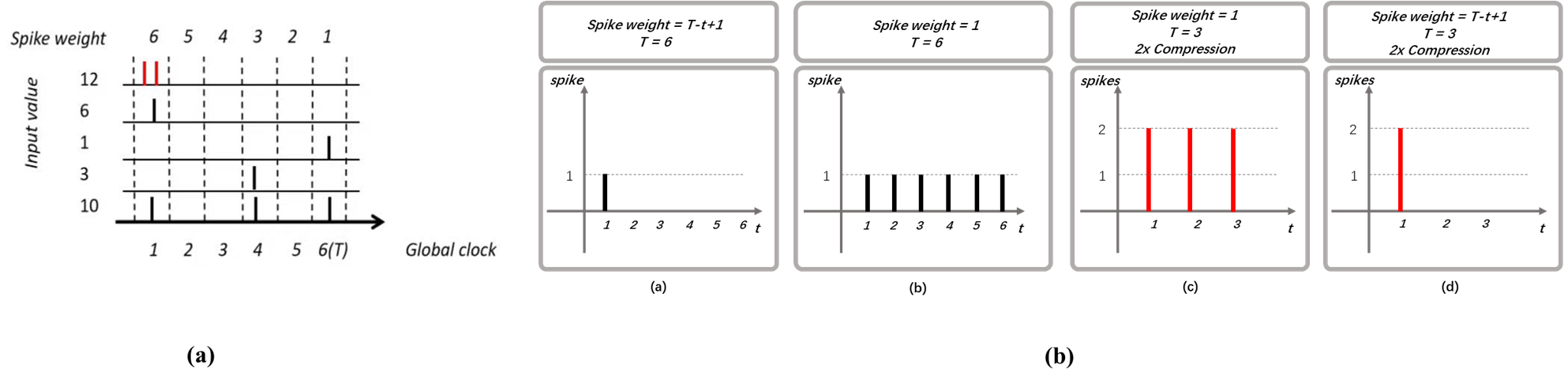}
    \caption{Spike-time-dependent integrated (STDI) coding. (a) shows the correspondence between the input value and the moment of spike issuance. (b) illustrates the main process of information transfer after using STDI. Where the accumulation phase is in $t\in[-T,0]$. In the firing phase, the red dashed line is the variable threshold over time.}
    \label{fig:STDIcoding}
\end{figure*}

the TS scheme was employed in the present study, leading to the observation of two distinct phases of STDI which are the accumulation phase and the firing phase. In the accumulation phase, the input information is accumulated to the membrane potential according to the Eq. \ref{eq:uin of spike-time-dependent integrated coding method}.
During the firing phase, the threshold decrease according to the time. the accumulated membrane potential searches for the right firing time to release the spikes. Specifically, when the membrane potential is less than the spike threshold at the current moment, the spike is not fired and waits for the next moment. When the membrane potential is greater than the spike threshold at that moment, the spike fires and the membrane potential decreases according to the subtractive reset~\cite{rueckauer2017conversion}. Cycle this process until the membrane potential is lower than $V_{thr}$. 
The specific algorithm is shown in Algorithm \ref{algorithm:2}.
In the output layer, we need to decode the spike train to spiking ratio according to Eq. \ref{eq:fasttimedecoder}. Fig. \ref{fig:STDIcoding}(b) illustrates the whole process.

\begin{algorithm}
    \caption{Algorithm for STDI }
    \label{algorithm:2}
        \begin{algorithmic}
            \REQUIRE $T$, $x$
            \ENSURE $s$
                \STATE Initialize $V_{mem} $ to 0
                \STATE Initialize $s$ to 0\\
                \STATE phase 1:
                \FOR{$t$ = 1 \TO $T$}
                    \STATE $\tau(t) \gets T-t+1$
                    \STATE $V_{mem} \gets V_{mem}$ + $x$[$t$-1]*$\tau(t)$
                \ENDFOR
                \STATE phase 2:
                \IF{$V_{mem}\leq0$}  
                    \STATE $V_{mem} = 0$
                \ENDIF
                \FOR{$t$ = 1 \TO $T$}
                    \STATE $\tau(t) \gets T-t+1$
                    \STATE $s[t-1] \gets s[t-1] + \lfloor V_{mem} / \tau(t) \rfloor$\\
                    $V_{mem} \gets V_{mem}$ - $s$[$t-1$]*$\tau(t)$
                \ENDFOR
                \RETURN{$s$}
        \end{algorithmic}
\end{algorithm}

\begin{figure}[htbp]
    \centering
    \includegraphics[width=0.48\textwidth]{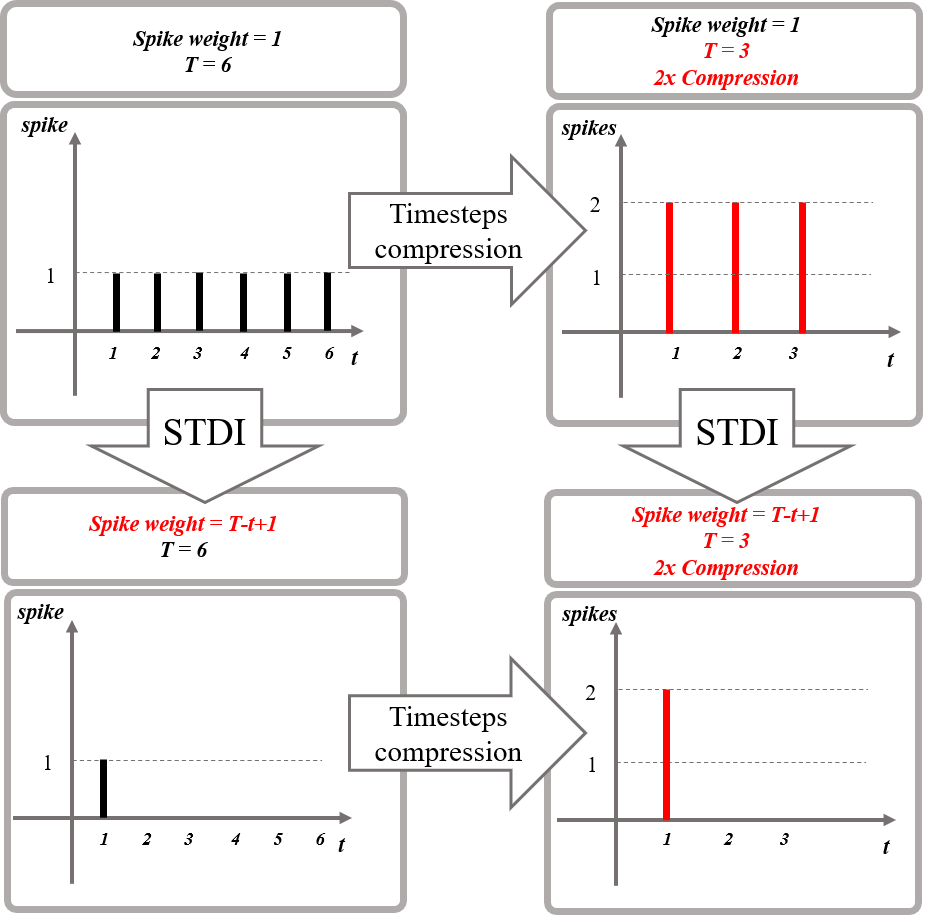}
    \caption{Timesteps compression and STDI. Here we give an example of the effect of applying the timesteps compression and STDI. Assuming timesteps $T$ is 6, frequency coding, timesteps compression, and STDI are used to express $spiking~ratio = 1$, respectively. Where the black vertical line is the binary spike and the red vertical line is the burst spike.}
    \label{fig:two methods scheme}
\end{figure}

 The advantage of applying STDI is that intercellular messaging can be accomplished with very few spikes. In combination with timesteps compression, we can express and convey information with fewer spikes and much lower timesteps. An example of this characteristic is given in Fig. \ref{fig:two methods scheme}. This example demonstrates that with the use of STDI and timesteps compression, information that would otherwise require six spikes can be expressed with only one binary or burst spike, with a correspondingly lower timesteps. Interestingly, STDI makes the spikes conform to the poisson distribution with or without the use of TS.

\subsection{SPPF Structure Conversion}
After comparing the YOLO series models, we chose the YOLOv5s as the backbone of our ANN to SNN conversion. Compared to the YOLOv3-tiny used by Spiking-YOLO, YOLOv5s has a deeper network and an optimized feature extraction structure such as FPN+PAN (i.e. SPPF), making it more interesting for development. 

\begin{figure*}[htbp]
    \centering
    \includegraphics[width=0.95\textwidth]{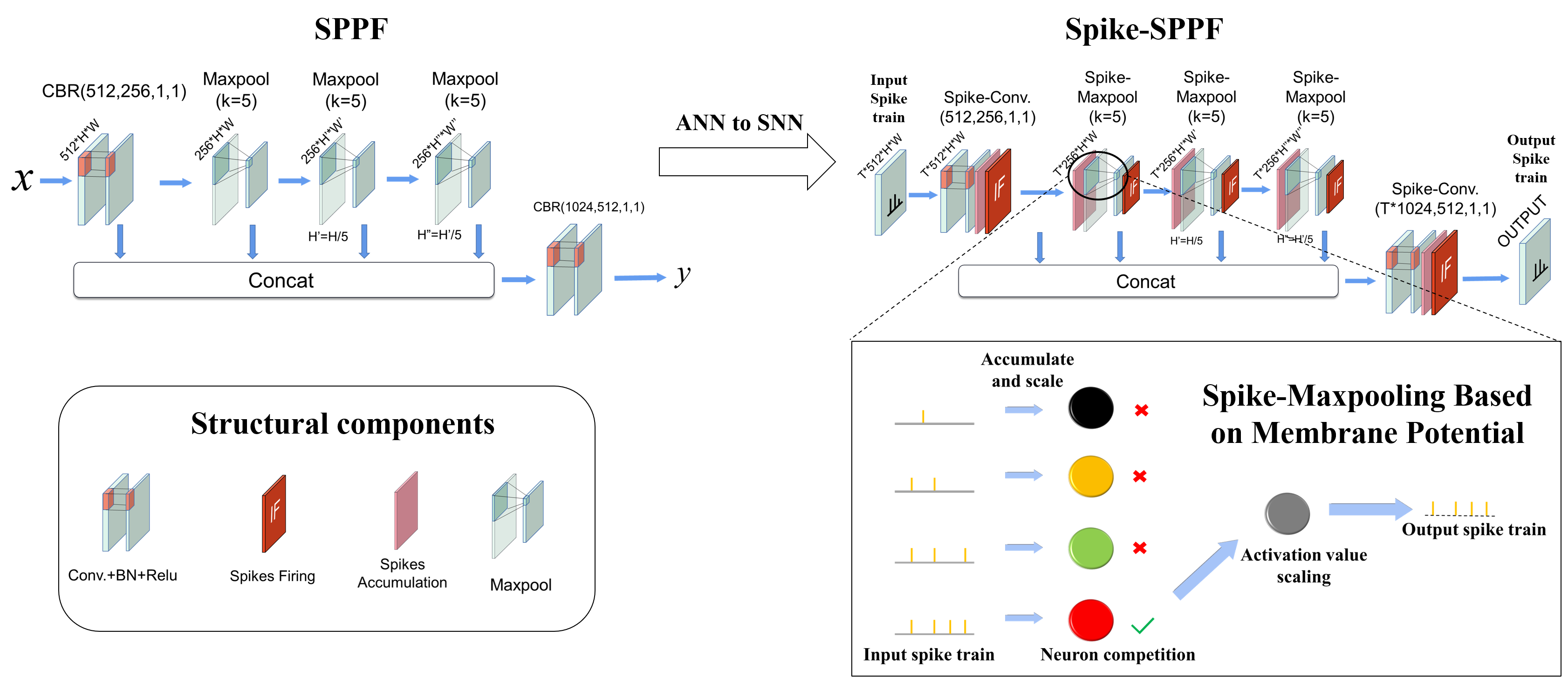}
    \caption{The comparison of SPPF and Spike-SPPF. The traditional SPPF (left figure) contains two CBR layers and three Maxpool layers with stride=1 and connects these layers through a concat operation. As a comparison, Spike-SPPF uses Spike-Conv. layer and Spike-Maxpooling based on membrane potential to achieve the same operation on the spike train. The extended diagram of the circle in the right figure shows the specific method of Spike-Maxpooling based on the membrane potential, which includes five stages.}
    \label{fig:structuremodify}
\end{figure*}

We first use ReLU as the activation function and then replace the Upsampling layer with a Convtranspose layer to form the ANN version of SUHD. Compared to the previous ANN to SNN works, the SUHD conversion work has an additional SPPF structure's conversion.
SPPF structure includes three Maxpool layers with $stride=1$, as shown in SPPF in Fig. \ref{fig:structuremodify}. According to the conversion method of YOLOv3-tiny in Spiking-YOLO~\cite{spiking-yolo}, we first set the SPPF layer directly to the CBR layer (Conv. + BN + ReLU).

We get significant performance improvements after using this structure. The results are shown in Sec. \ref{sec:Optimization of accuracy and speed with Different structures}. However, the SPPF structure of the ANN version of SUHD led to an increase of approximately 3\% mAP, while the conversion method of the SPPF structure can be extended to the whole YOLO series and even more SNN based deep learning models, thus facilitating the conversion of more efficient and complex structures. Therefore, we perform the conversion of the SPPF structure. For achieving fast Maxpool operations with any stride in deep SNN models, we propose a Spike-Maxpooling mechanism based on membrane potentials.\\
\subsubsection{Spike-Maxpooling According to Membrane Potential}\label{sec:Pooling neurons according to membrane voltage}The existing Maxpool methods in SNNs mainly include two kinds: one is the "winner take all" ~\cite{Spiking-maxpooling,hu2016max}, that is, the pooling neurons accept all the previous spikes and add them. The pooling layer value obtained by this pooling method is often greater than the real value. The other one is to calculate the number of spikes of maximum firing rate neurons directly or indirectly. These methods are complicated, especially in the case of compressed timesteps and STDI.
Therefore, we proposed a Spike-Maxpooling method based on membrane potential, which allows for the Maxpool of SNNs at any stride in a simple way.

As shown in Spike-Maxpooling Based on Membrane Potential in Fig. \ref{fig:structuremodify}, the main process has five phases:

\begin{enumerate}
\item{\textbf{Accumulation phase}:} The input spiking train is decoded to obtain the input voltage $U_{in}$ and accumulated to the membrane.

\item{\textbf{Restore Normalization}:} Restore the activation values by Eq. \ref{eq:restore normalization}.
\begin{equation}
    \label{eq:restore normalization}
    V_r=max_{in}*U_{in},
\end{equation}

where $max_{in}$ denotes the maximum sample activation value of the input of this channel in ANN, and the $V_r$ denotes the activation value of this channel after restoring normalization in SNN. The purpose of this phase is to eliminate errors due to integer and floating point calculations.

\item{\textbf{Pooling phase}:} We name the region where the neurons that are involved in $V_r$ competition as a candidate region, as shown within the circle in the Fig. \ref{fig:structuremodify}. Neurons in candidate regions undergo $V_r$ competition. The winner's $V_r$ is passed on.

\item{\textbf{Re-normalization phase}:} Normalize $V_r$ value to output voltage as $U_{out}$ by Eq. \ref{eq:normalize membrain potential}.
\begin{equation}
    \label{eq:normalize membrain potential}
    U_{out}=\frac{V_r}{max_{out}},
\end{equation}

where $max_{out}$ denotes the sample maximum activation value of the output of this channel in ANN.

\item{\textbf{Firing phase}:} Encoding the $U_{out}$ to spike train and release it. \\
\end{enumerate}
\subsubsection{Spike-SPPF}
Based on the above work, we completed the conversion of the Spike-SPPF structure, the exact structure is shown in Spike-SPPF in Fig. \ref{fig:structuremodify}.
 The results, as shown in Sec. \ref{sec:Optimization of accuracy and speed with Different structures}, demonstrate that the accuracy of the SNN model improves by about 3\% after using Spike-SPPF.
\section{Experiments}
\label{sec:Experiments}
\subsection{Experiments Setup}
\label{sec:Experiments setup}
Different comparative experiments were set up to verify the effectiveness of different structures or methods. The model is the ANN/SNN version of the SUHD model. The initial conversion method (SNN base code) is frequency coding + TS. The membrane potential was initialized with a value of 0.5 with reference to the \cite{initialmemvoltage}. The whole experiment is based on Intel Core i7-8700K CPU or NVIDIA RTX2080Ti GPU with CUDA 10.1. The datasets used in this work are PASCAL VOC~\cite{everingham2009pascalVOC} and MS COCO~\cite{lin2014MSCOCO}, where the PASCAL VOC dataset consists of three parts, train (2007+2012), val (2007+2012), and test (2007), including 8218, 8333, and 4952 images respectively. Specifically, we used train (2007+2012) as the training set and test (2007) as the validation set. The COCO dataset consists of two parts, train2017 and val2017, with train2017 including 118,287 images and val2017 including 5,000 images. We used train2017 as the training set and val2017 as the evaluation set. Unless explicitly specified, the metric utilized to assess the accuracy of experiments is the mAP@0.5.
\subsection{Ablation Study of Spike-SPPF}
\label{sec:Optimization of accuracy and speed with Different structures}

In order to demonstrate the impact of Spike-SPPF on improving the accuracy upper bound, an ablation study was conducted with and without Spike-SPPF. To ensure impartiality and proper operation of the model, the SPPF component was replaced with a CBR (Conv. + BN + ReLU) layer in ANN of the model without Spike-SPPF. The results of the ablation study are shown in Tab. \ref{tb:performence of different structures}.

\begin{table}[htbp]
    \centering
    \caption{The effect of Spike-SPPF on accuracy}
    \fontsize{10}{12}\selectfont
      \begin{tabular}{c|cc}
        \toprule
        \multirow{2}{1cm}{Structure} & \multicolumn{2}{c}{Acc.(mAP@0.5\%)} \\
        &VOC&COCO\\
        \midrule
            {w/o Spike-SPPF} &72.9&52.4\\
            {w Spike-SPPF} &75.3&54.6\\
        \bottomrule
    \end{tabular}
\label{tb:performence of different structures}
\end{table}
The results demonstrate that the Spike-SPPF can improve accuracy by approximately 2.3\%. In particular, the accuracy loss of the ANN to SNN conversion here does not exceed 0.2\%, which verifies that our proposed Spike-Maxpooling has the ability to losslessly convert the Maxpool layer with $stride=1$. 
\subsection{Optimization of Accuracy and Speed Ablation Experiments}
\label{sec:Optimization of Accuracy with Different Methods}
Our proposed methods aim to achieve a faster and more precise SNN-based object detection model within limited timesteps. In order to evaluate the effectiveness of these methods, we conducted ablation experiments to measure the impact on speed and accuracy improvements.
\begin{table*}
\centering
\caption{Accuracy and speed improvement by different methods on PASCAL VOC}
\fontsize{10}{12}\selectfont
\begin{tabular}{ccccc}
    \toprule
      \multirow{2}{*}{Methods}&\multirow{2}{*}{\makebox[0.1\textwidth]{\makecell{Timesteps}}}&\multirow{2}{*}{\makebox[0.2\textwidth]{\makecell{mAP@0.5}}}&\multicolumn{2}{c}{\makebox[0.13\textwidth]{\makecell{Speed(ms/frame)}}} \\
      &&&CPU&GPU \\
      \midrule
        {ANN}&-&{75.3}&-&-\\ 
      \midrule
        {Basecode}&{64}&{72.5}&{6639.2}&{720} \\
      \midrule
        {16x Compression}&4&72.5&867.3&185.8 \\
      \midrule
        {16x Compression + STDI}&\textbf{4}&\textbf{75.3}&541&\textbf{152.3} \\
    \bottomrule
\end{tabular}  
\label{tb:1}
\end{table*}
\begin{table*}
\centering
\caption{Accuracy and speed improvement by different methods on MS COCO}
\fontsize{10}{12}\selectfont
\begin{tabular}{ccccc}
    \toprule
      \multirow{2}{*}{Methods}&\multirow{2}{*}{\makebox[0.1\textwidth]{\makecell{Timesteps}}}&\multirow{2}{*}{\makebox[0.2\textwidth]{\makecell{mAP@0.5}}}&\multicolumn{2}{c}{\makebox[0.13\textwidth]{\makecell{Speed(ms/frame)}}} \\
      &&&CPU&GPU \\
      \midrule
        {ANN}&-&{54.8}&-&-\\ 
      \midrule
        {Basecode}&{64}&{53.6}&{6562}&{770} \\
      \midrule
        {16x Compression}&4&53.6&882.6&185.9 \\
      \midrule
        {16x Compression + STDI}&\textbf{4}&\textbf{54.6}&{552}&\textbf{155.9} \\
    \bottomrule
\end{tabular}  
\label{tb:2}
\end{table*}
The results are presented in Tab. \ref{tb:1} and \ref{tb:2}. In these experiments, we used a base code with uncompressed timesteps and frequency encoding, then gradually increased the compression scale and applied STDI. The initial timesteps were set to 64, and the model achieved an mAP of 72.5\% and 53.6\% on the PASCAL VOC and MS COCO datasets, respectively, with a processing speed of over 6500 (CPU)/720 (GPU) ms per frame.

Increasing the compression scale to 16 and reducing the compressed timesteps to 4 significantly improved the processing speed without compromising accuracy, indicating the effectiveness of timesteps compression.

Changing the encoding method to STDI further improved the mAP by 2.8\% and 1\% on the PASCAL VOC and MS COCO datasets, respectively, and reduced the inference time by about 38\% (CPU)/17\% (GPU), confirming the effectiveness of STDI in reducing the conversion error and improving the speed of object detection.

Finally, we increased the compression scale to 64 with compressed timesteps of 1, without reducing detection accuracy. This allowed us to achieve the fastest detection speeds of 189.4ms/frame and 189.5ms/frame (CPU) on PASCAL VOC and MS COCO datasets, respectively. We also conducted the same experiment on the GPU and achieved a speed of 90.1ms/frame at 64x compression.
\subsection{Compared with the-State-of-the-Arts}
\begin{figure}[htbp]
    \centering
    \includegraphics[width=0.48\textwidth]{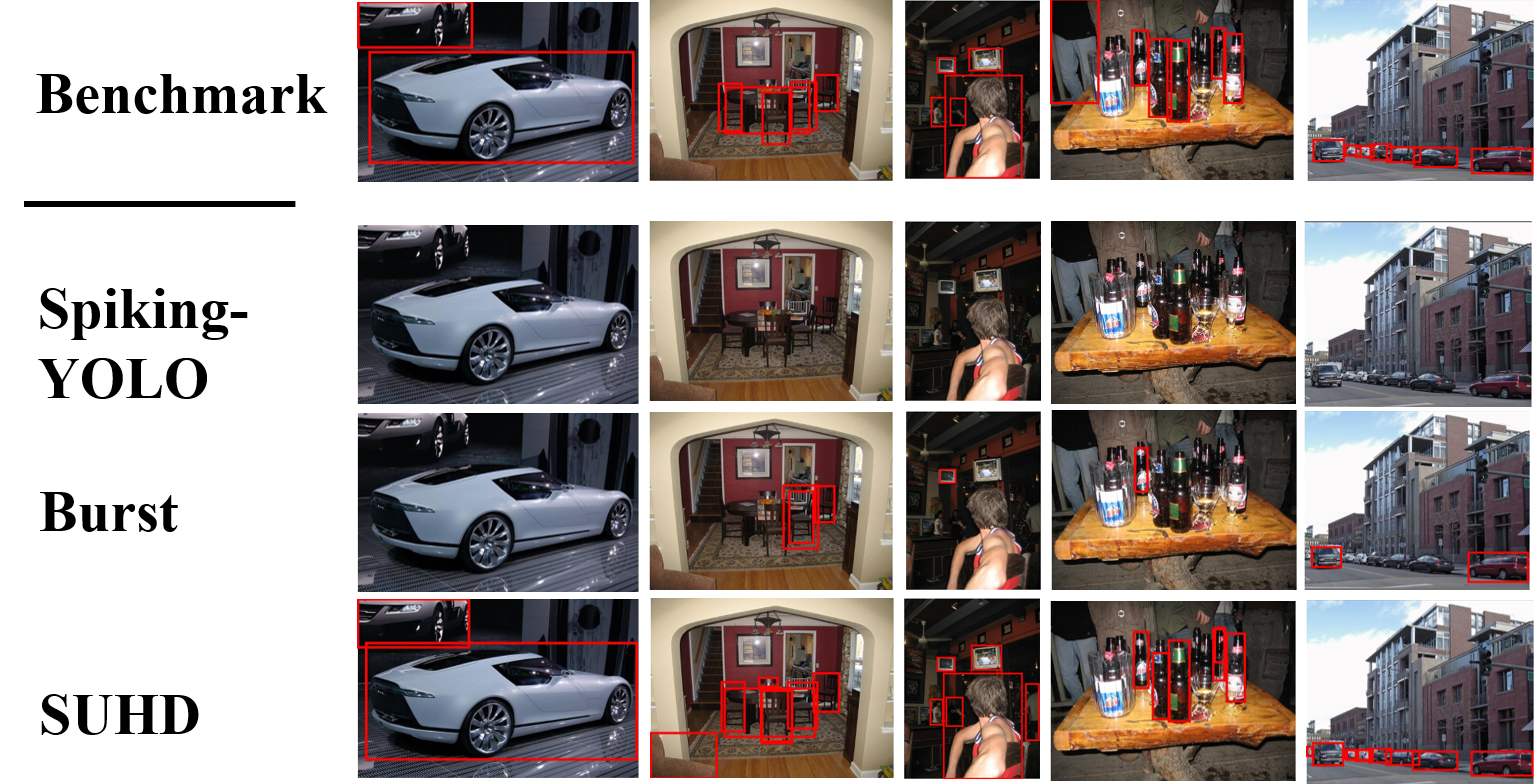}
    \caption{Detection results using different methods. Our work achieved the best results in detection.}
    \label{fig:hahaha structure}
\end{figure}

We compared the performance of SUHD with that of other methods on the PASCAL VOC and MS COCO datasets respectively. The results are shown in Tab. \ref{tab:VOC data set mAP situation comparison}, Tab. \ref{tab:COCO data set mAP situation comparison} and Fig. \ref{fig:hahaha structure}, where Burst refers to applying burst spikes to prevent harm caused by truncation errors~\cite{burst}. The data in tables are taken from the corresponding papers~\cite{burst,spiking-yolo,kim2020towards}.
\label{sec:results}
\begin{table*}[htbp]
    \centering
    \caption{Comparison with other works under the PASCAL VOC dataset}
    \fontsize{10}{12}\selectfont
        \begin{tabular}{ccc}
            \toprule
                \makebox[0.26\textwidth]{\makecell{Model}}&{\begin{tabular}{c}mAP@0.5\\(PASCAL VOC)\end{tabular}}&\makebox[0.26\textwidth]{\makecell{Timesteps}}\\
            \midrule
                \makebox[0.26\textwidth]{\makecell{Spiking-YOLO\\(Kim et al., 2020)~\cite{spiking-yolo}}}&\makebox[0.26\textwidth]{\makecell{51.83}}&\makebox[0.26\textwidth]{\makecell{8000}}\\
            \hline
                \makebox[0.26\textwidth]{\makecell{\begin{tabular}{c}
                    Vthfast+Vthacc\\(Kim et al., 2020)~\cite{kim2020towards}
                \end{tabular}}}
                &\makebox[0.26\textwidth]{46.66}&\makebox[0.26\textwidth]{500}\\
            \hline 
                \makebox[0.26\textwidth]{
                \begin{tabular}{c}
                    Burst+MLIpooling+SpiCalib\\(Yi Zeng et al., 2022)~\cite{li2022spike}
                \end{tabular}}&\makebox[0.26\textwidth]{75.21}&\makebox[0.26\textwidth]{512}\\
            \hline
            \begin{tabular}{c}
                \textbf{SUHD}\\ \textbf{(our model)}
            \end{tabular}&\makebox[0.26\textwidth]{\textbf{75.3}}&\makebox[0.26\textwidth]{\makecell{\textbf{4}}}\\
            \bottomrule         
            \end{tabular}
    \label{tab:VOC data set mAP situation comparison}
\end{table*}
\begin{table*}[htbp]
    \centering
    \caption{Comparison with other works under the MS COCO dataset}
    \fontsize{10}{12}\selectfont
        \begin{tabular}{ccc}
            \toprule
                \makebox[0.26\textwidth]{\makecell{Model}}&\makebox[0.26\textwidth]{\makecell{\begin{tabular}{c}mAP@0.5\\(MS COCO)\end{tabular}}}&\makebox[0.26\textwidth]{\makecell{Timesteps}}\\
            \midrule
                \makebox[0.26\textwidth]{\makecell{Spiking-YOLO\\(Kim et al., 2020)~\cite{spiking-yolo}}}&\makebox[0.26\textwidth]{\makecell{25}}&\makebox[0.26\textwidth]{\makecell{3000}}\\
            \hline 
                \makebox[0.26\textwidth]{\makecell{\begin{tabular}{c}
                    Vthfast+Vthacc\\(Kim et al., 2020)~\cite{kim2020towards}
                \end{tabular}}}&\makebox[0.26\textwidth]{\makecell{21.05}}&\makebox[0.26\textwidth]{\makecell{500}}\\
            \hline
             \makebox[0.26\textwidth]{\makecell{\begin{tabular}{c}
                   FSHNN\\(Chakraborty et al., 2021)~\cite{chakraborty2021fully}
                \end{tabular}}}&\makebox[0.26\textwidth]{\makecell{42.6}}&\makebox[0.26\textwidth]{\makecell{300}}\\
            \hline
            \makebox[0.26\textwidth]{\makecell{\begin{tabular}{c}
                    Burst+MLIpooling+SpiCalib\\(Yi Zeng et al., 2022)~\cite{li2022spike}
                \end{tabular}}}&\makebox[0.26\textwidth]{\makecell{45.42}}&\makebox[0.26\textwidth]{\makecell{512}}\\
            \hline
            \begin{tabular}{c}
                \textbf{SUHD}\\ \textbf{(our model)}
            \end{tabular}&\makebox[0.26\textwidth]{\makecell{\textbf{54.6}}}&\makebox[0.26\textwidth]{\makecell{\textbf{4}}}\\
            \bottomrule   
            \end{tabular} 
    \label{tab:COCO data set mAP situation comparison}
\end{table*}
In comparison with the current advanced spiking object detection, our methods have achieved the optimal result in terms of speed, precision, and timesteps. As shown in the Tab. \ref{tab:VOC data set mAP situation comparison}, with the PASCAL VOC dataset, our proposed method achieves an accuracy improvement of about 23\% using 2000x fewer timesteps compared to Spiking-YOLO. Compared to the Burst + MLIpooling + SpiCalib method, we achieve almost the same accuracy using 128x fewer timesteps. The results under the MS COCO dataset are shown in Tab. \ref{tab:COCO data set mAP situation comparison}. Compared to the Spiking-YOLO and Burst + MLIpooling + SpiCalib methods, our method achieves about 30\% and 9\% improvement in accuracy using 750x and 128x fewer timesteps, respectively. Compared to FSHNN, we achieve a 12\% accuracy improvement using 75x fewer timesteps.
\subsection{Model Robustness}\label{sec:Model Robustness}
The noise immunity of the model is also one of the main indicators of model performance. To verify the stability of the model, we evaluate the performance of the model under adverse conditions using standard additive gaussian white noise with signal-to-noise ratios (SNR) of 15dB and 30dB respectively. The FSHNN~\cite{chakraborty2021fully} model was also added for comparison.
The results are shown in Tab. \ref{tab:modelrobust}.
\begin{table}[htbp]
    \centering
    \caption{Comparison of the robustness of different models under different input noise}
    \begin{tabular}{c|cc|cc|cc}
        \toprule
        \multirow{2}{0.6cm}{SNR}&\multicolumn{2}{c}{ANN}&\multicolumn{2}{c}{SUHD}&\multicolumn{2}{c}{FSHNN}\\
        &VOC&COCO&VOC&COCO&VOC&COCO\\
        \midrule
         Clean signal& 75.3& 54.8& 75.3& 54.6&-&42.7  \\
         30dB& 73.5& 54.3& 73.1& 54.1&-&41.2\\
         15dB& 37.7& 40.1& 37.2& 39.9&-&33.0\\
        \bottomrule
    \end{tabular}
    \label{tab:modelrobust}
\end{table}
\begin{figure}
    \centering
    \includegraphics[width=0.48\textwidth]{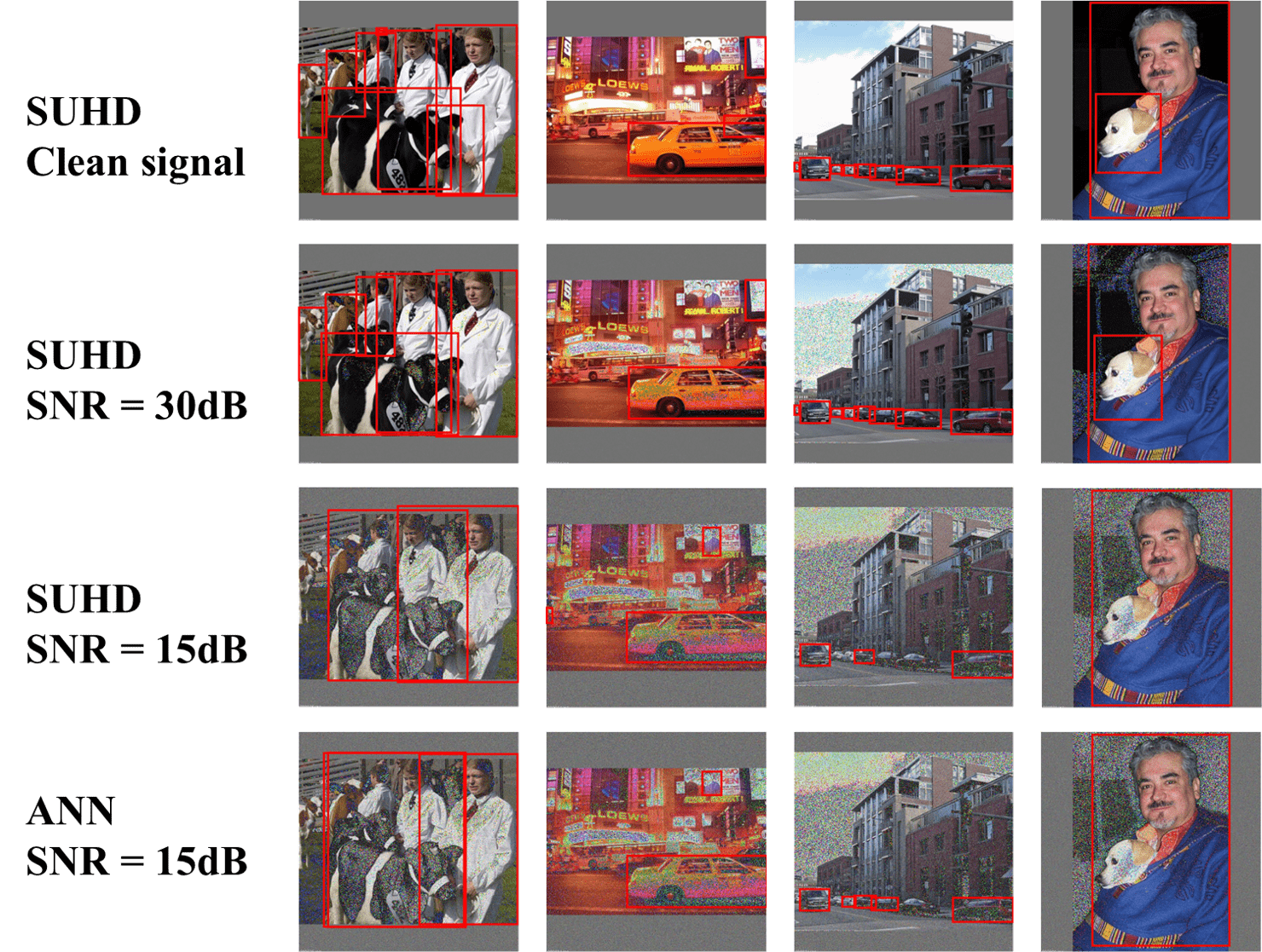}
    \caption{Performance of the models at different SNR. The first three rows show the detection results using the SUHD model and the last row shows the detection results using the ANN model.}
    \label{fig:Model Robustness}
\end{figure}
When SNR = 30 dB, we can see that our model has almost no loss($\leq 2.2\%$). When SNR = 15 dB, accuracy suffered a more serious decline. However, the model still has an advantage of approximately 6.9\% over the FSHNN. At the same time, the noise immunity of the SUHD model is not inferior to that of the ANN version of SUHD at high noise levels($loss \leq 0.5\%$), as shown in Tab. \ref{tab:modelrobust} and Fig. \ref{fig:Model Robustness}. This suggests that our proposed methods do not have an excessive impact on the robustness of the model. Thus, our proposed model has satisfactory performance in terms of robustness.
\subsection{Energy Efficiency}
Energy consumption is an important metric for measuring the cost of model inference. Referring to other works~\cite{spiking-yolo,9746375,chakraborty2021fully}, the generic index $FLOPs$ is used to measure the complexity of the model. The $energy$ denotes the energy cost of models. As proposed in Horowitz et al. ~\cite{Horowitz,spiking-yolo}, the energy cost of one operation is 4.6pJ(FLOAT32 MAC)/0.9pJ(FLOAT32 AC).
To measure SNN's energy consumption more rationally, we use the average FLOPs based on the conversion sample dataset. Specifically, we define SNN FLOPs as:
\begin{equation}
    \label{eq:efficient}
    FLOPs=\sum_{l=1}^n \sum_{t=1}^T s^l(t) ~+~\sum p_{in},
\end{equation}

where $n$ denotes the sum of layers of the model, and the $\sum p_{in}$ represents the sum of pixels of the input figures. Considering that the model accepts analog inputs, we define the input layer operations as MAC operations. The other operations are caused by spikes as floating-point AC operations. In contrast to spiking-yolo, the SUHD model uses YOLOv5s as its backbone. Therefore, our evaluation is between YOLOv5s and SUHD. The energy efficiency profile of YOLOv5s is derived from official data.
Based on this method, we compared the YOLOv5s and the SUHD's energy costs.

\begin{table}[htbp]
    \centering
    \caption{Energy efficiency}
    \fontsize{10}{12}\selectfont
        \begin{tabular}{cccc}
        \toprule
         & FLOPs& Power(w)& Energy(J) \\
        \midrule
         YOLOv5s& 1.67E+10& 10.97& 7.7E-02\\
         SUHD& 2.05E+08& 4.1E-03& 3.7E-04 \\
        \bottomrule
    \end{tabular}
    \label{tb:efficient results}
\end{table}
The results are shown in Tab. \ref{tb:efficient results}, which illustrate that the SUHD model is at least 200 times more energy efficient than the YOLOv5s model.

\subsection{Algorithm Deployment and Evaluation in Robot Platform }
To evaluate the ability of the model to detect objects in dynamic scenarios and the performance of the model on mobile platforms, in this section we deploy the algorithm to a mobile robot platform and test it.
\subsubsection{Algorithm Deployment}
\begin{figure}
    \centering
    \includegraphics[width=0.38\textwidth]{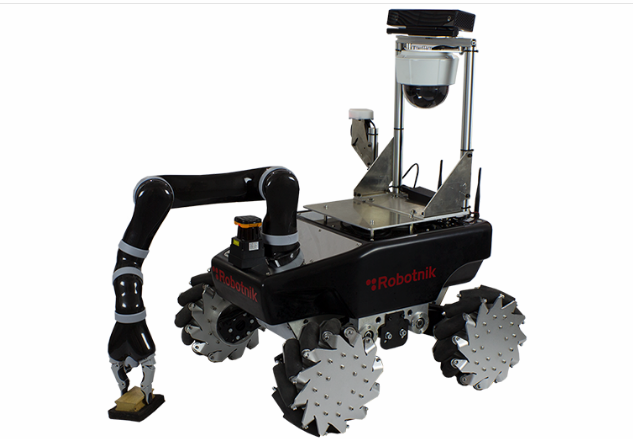}
    \caption{Summit-xl mobile robot platform.}
    \label{fig:summit-xl}
\end{figure}
We use the summit-xl mobile robot, as shown in Fig. \ref{fig:summit-xl}. Summit-xl is equipped with an Intel Core i3-9100 CPU, 7.16GB RAM, four independent drive wheels, and an Axis gimbal camera. The deployment process consists of three main steps: We first built the python algorithm runtime environment based on ubuntu 16.04 on x86 architecture and deployed the algorithm to the robot. The video signal is then acquired using the gimbal camera. The video signal is in h265 format and is hard-decoded. Finally, the decoded video image stream is fed into the model in real time for object detection.
\subsubsection{Performance Evaluation in Dynamic Scenarios}
We produced two video datasets. Both datasets contain one video each.\\
\textbf{Dataset A} was sampled from the robot gimbal camera at 6 fps. Three motion speeds, fast, medium, and slow were included throughout the sampling process. At the same time, both the robot and the object can move. This dataset is therefore comprehensive and can accurately assess the performance of the algorithm in dynamic scenes. The entire dataset contains 920 frames with a total of 4140 labels in 9 categories, with an average of 4.5 labels per frame.\\
\textbf{Dataset B} was sampled from a fixed viewpoint and contains 50 frames in 3 categories, with a total of 219 labels, averaging 4.38 labels per frame. The overall speed of object movement within the frames is slow.

\begin{figure*}[t]
    \centering
    \includegraphics[width=1\textwidth]{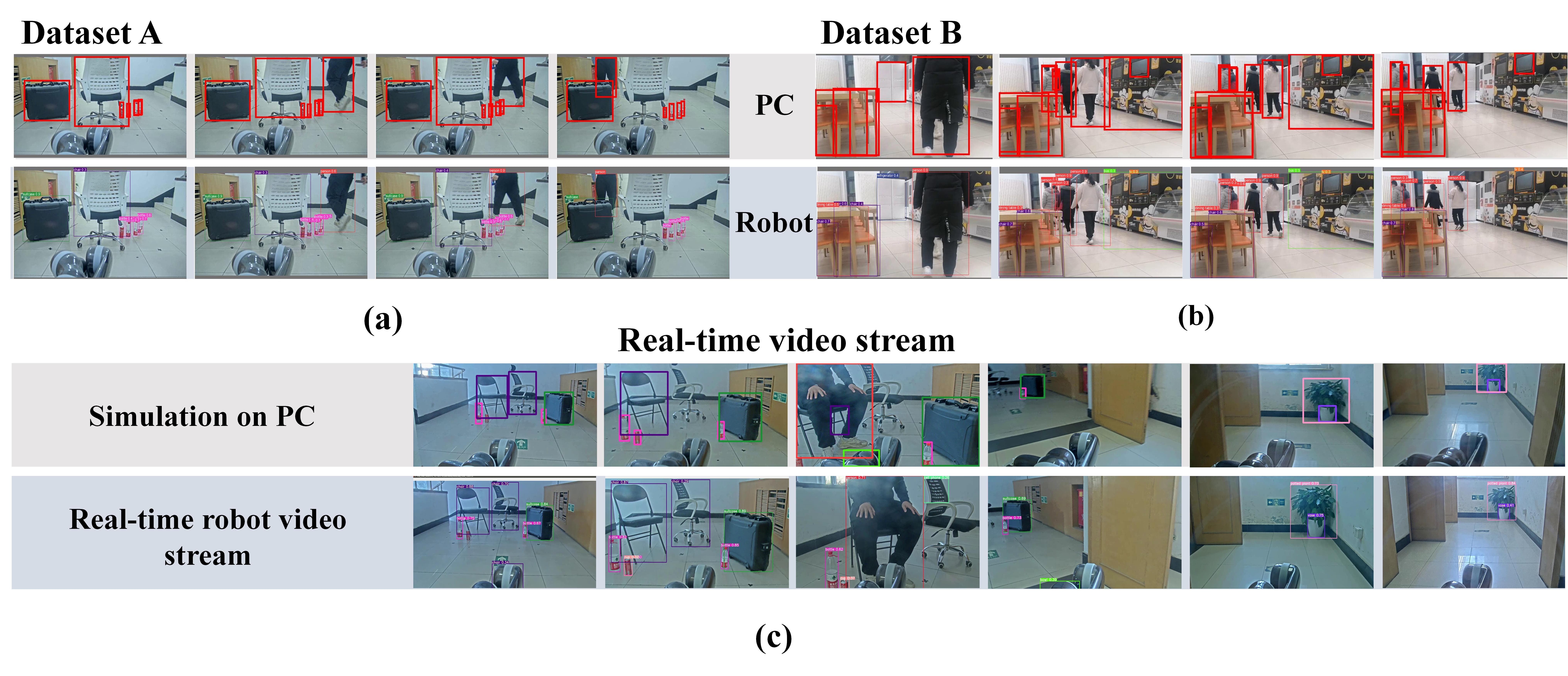}
    \caption{Performance evaluation in dynamic scenarios. (a) and (b) show detection results for Dataset A and Dataset B, respectively. The top row shows PC results and the bottom row shows robot results. (c) presents detection results for video streaming, with the top row showing the results for the simulated real-time video streaming on the PC and the bottom row showing the results for the real-time robot video streaming on the robot.}
    \label{fig:final}
\end{figure*}

First, we train the ANN model using the COCO dataset and then convert it into an SNN model. In the second step, we deploy the SNN models to the PC and mobile robot described at the beginning of this section, respectively. In the third step, we use datasets A and B to compute the mAP to obtain the object detection capability of the model in dynamic scenes. In the fourth step, we test the object recognition capability of the model deployed on the robot using a real-time robot video streaming. At the same time, we record video data in the same scene and then move the video to the PC to simulate object detection in a real-time dynamic scene.
\begin{table}[htbp]
    \centering
    \caption{Detection accuracy on PC and robot side}
    \fontsize{10}{12}\selectfont
      \begin{tabular}{c|cc|c}
        \toprule
        \multirow{2}{1cm}{Device} & \multicolumn{2}{c}{mAP@0.5} & {Recognition accuracy}\\
        &Dataset A&Dataset B&Real-time signal\\
        \midrule
            {The PC} &69.8&92.5&74.6\\
            {Robot} &69.8&92.5&71.9\\
        \bottomrule
    \end{tabular}
\label{tb:alogrithmdeployment}
\end{table}
The results were shown in Tab. \ref{tb:alogrithmdeployment}, where the evaluation indicator of Dataset A and Dataset B is mAP@0.5 for detection accuracy, and the evaluation indicator of the real-time robot video streaming is recognition accuracy, which is defined as follows:
\begin{equation}
    Acc = \frac{TP + TN}{P+N},
\end{equation}

where $Acc$ is recognition accuracy, $TP+TN$ is the sum of the number of objects correctly classified, and $P+N$ is the sum of the number of all objects. The results of Dataset A and B show that the model deployed on the robot produced no additional losses. This result indicates that our algorithm has strong repeatability. Due to the use of moving observation points in dataset A, there is a more significant blurring of some of the frames, which results in an mAP of 69.8\%. Dataset B uses a fixed viewpoint and therefore achieves an mAP of 92.5\%. Fig. \ref{fig:final}(a) (b) show the difference in performance between the model on the PC and the model deployed to the mobile robot. In addition, we examined the object detection of the algorithmic model on the robot side under real-time robot video streaming and compared its performance with that of the PC side model under the same dynamic scene video. The results are shown in Fig. \ref{fig:final}(c) and the Real-time signal of Tab. \ref{tb:alogrithmdeployment}. The two results differ by only 2.7\%. The reason for the error is that the robot has less computing ability than the PC, resulting in greater frame drops and blurring of the live video signal it receives, thus reducing recognition precision. This result suggests that the detection performance of the model is not overly compromised by deployment to the robot side.


\section{Conclusion}
This paper presents three main methods for optimizing the conversion accuracy and run speed of SNNs. The first method is timesteps compression, which significantly reduces the required timesteps to the minimal one timestep during lossless conversion. The second method is STDI, which improves the efficiency of SNN inference by an average of 38\% on the CPU and 17\% on the GPU by increasing the information capacity of a single spike. The third approach is a Spike-Maxpooling mechanism based on membrane potential, which facilitates the lossless conversion of complex and efficient structures in ANNs.


By combining timesteps compression and STDI, we achieved results that are 34 times faster on the CPU than initial conversion method (frequency coding + TS). We have further built the spike-based high-performance object detection SUHD, which can run on mobile platforms with comprehensive performance up to the state-of-the-art. Especially, SUHD is currently the deepest object detection model that can achieve a lossless ANN to SNN conversion with an ultra low timesteps. Our methods have shown competitive results and have significant potential for improving the efficiency and accuracy of SNNs, while offering the possibility to solve the problem of deploying SNN models on the mobile terminal. We believe that our methods can contribute to the large-scale popularisation of SNNs application in the future.

\bibliographystyle{IEEEtran}
\bibliography{main}{}

\begin{IEEEbiography}[{\includegraphics[width=1in,height=1.25in,clip,keepaspectratio]{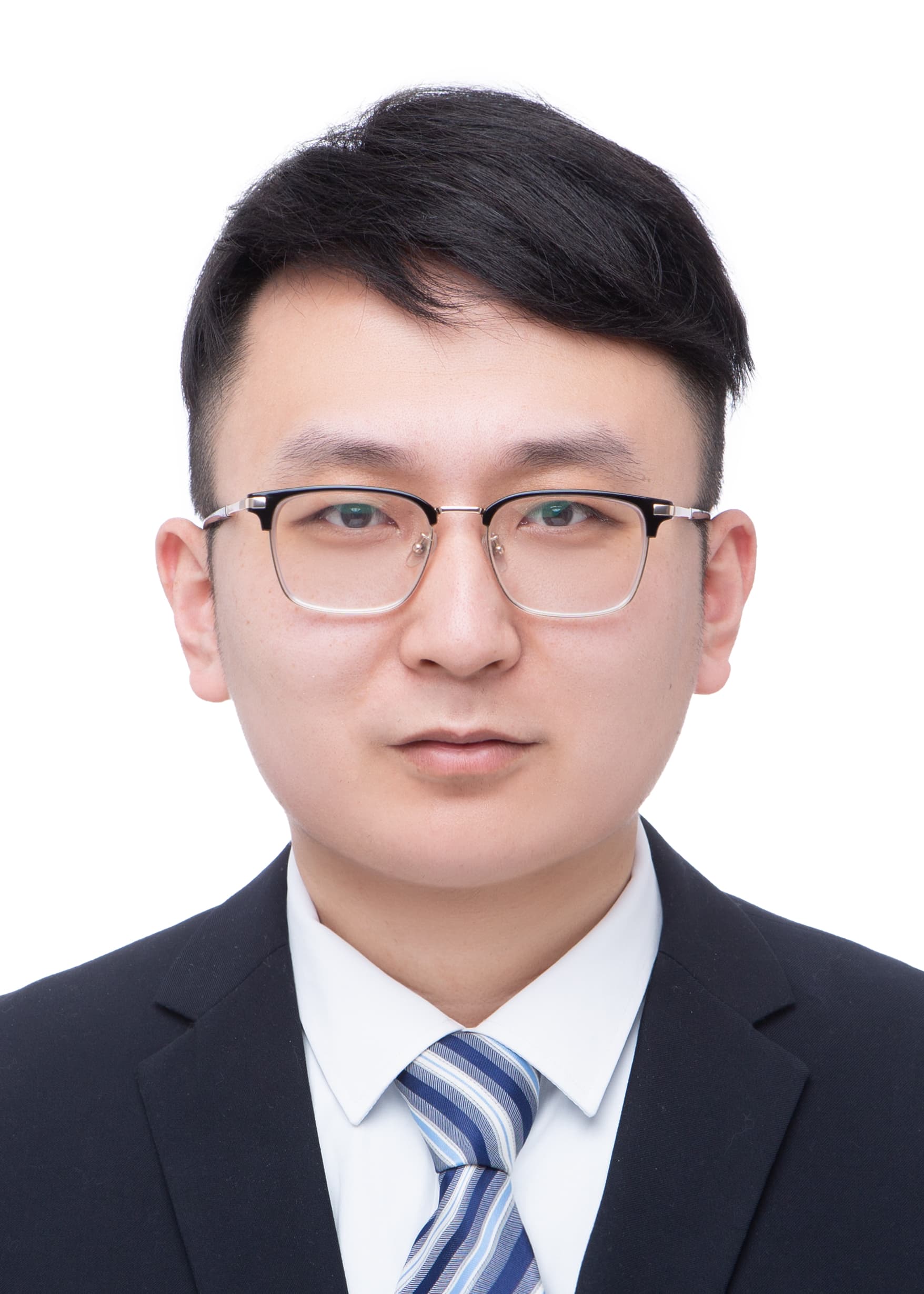}}]{Jinye Qu}
received the B.S. degree from Chongqing Jiaotong University, Chongqing, China in 2018. He is now pursuing the master’s degree at School of Artificial Intelligence, University of Chinese Academy of Sciences, Beijing, China. His research interests include brain-inspired intelligence, object detection, and object tracking.\end{IEEEbiography}

\begin{IEEEbiography}[{\includegraphics[width=1in,height=1.25in,clip,keepaspectratio]{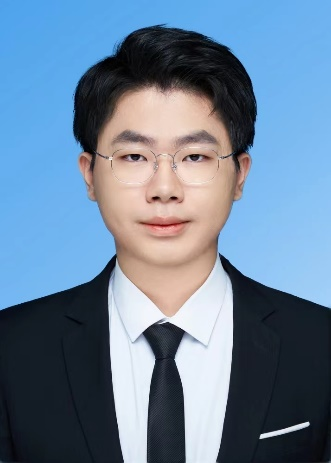}}]{Zeyu Gao}
received the B.S. degree in Intelligent Vehicle Engineering with School of Automotive Engineering, Harbin Institute of Technology, China in 2023. He is currently working toward the M.S. degree in pattern recognition and intelligent system at Institute of Automation, Chinese Academy of Science, China. His research interests include autonomous driving, reinforcement learning, and brain-inspired intelligence.\end{IEEEbiography}

\begin{IEEEbiography}[{\includegraphics[width=1in,height=1.0in,clip,keepaspectratio]{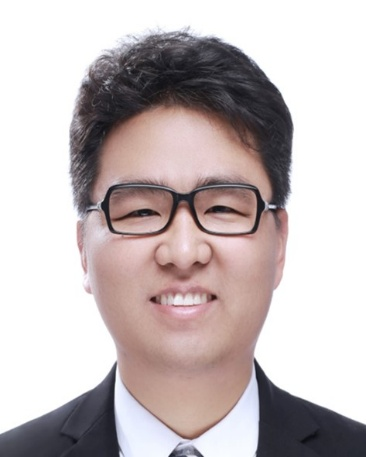}}]{Tielin Zhang} received the Ph.D. degree from the Institute of Automation Chinese Academy of Sciences, Beijing, China, in 2016. He is an Associate Professor in the Laboratory of Cognition and Decision Intelligence for Complex Systems, Institute of Automation, Chinese Academy of Sciences. His current interests include theoretical research on neural dynamics and Spiking Neural Networks.
\end{IEEEbiography}

\begin{IEEEbiography}[{\includegraphics[width=1in,height=1.25in,clip,keepaspectratio]{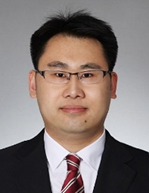}}]{Yanfeng Lu}
 (Member, IEEE) received his B.S. degree in Automation from the Harbin Institute of Technology, China in 2010, and his Ph.D. degree from Korea University, Republic of Korea in 2015. He is currently an Associate Professor with the State Key Laboratory of Multimodal Artificial Intelligence Systems, Institute of Automation, Chinese Academy of Sciences, Beijing, China. His research interests include brain-inspired computing, computer vision, robot vision, and machine learning.\end{IEEEbiography}

\begin{IEEEbiography}[{\includegraphics[width=1in,height=1.25in,clip,keepaspectratio]{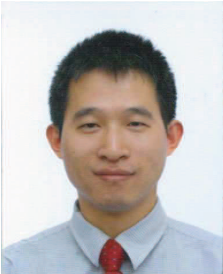}}]{Huajin Tang}
 (Member, IEEE) received the B.Eng. degree from Zhejiang University, China in 1998, received the M.Eng. degree from Shanghai Jiao Tong University, China in 2001, and received the Ph.D. degree from the National University of Singapore, in 2005. He is currently a professor at Zhejiang University, China. His research work on Brain GPS has been reported by MIT Technology Review in 2015. He received the 2016 IEEE Outstanding TNNLS Paper Award. His current research interests include neuromorphic computing, neuromorphic hardware and cognitive systems, robotic cognition, etc. Dr. Tang is the Editor-in-Chief of IEEE Transactions on Cognitive and Developmental Systems and Associate Editor of IEEE Transactions on Neural Networks and Learning Systems, and Frontiers in Neuromorphic Engineering. He was the Program Chair of the 6th and 7th IEEE CIS-RAM, and Chair of 2016 and 2017 IEEE Symposium on Neuromorphic Cognitive Computing.\end{IEEEbiography}

 \begin{IEEEbiography}[{\includegraphics[width=1in,height=1.25in,clip,keepaspectratio]{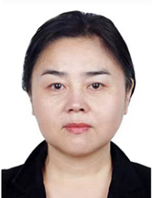}}]{Hong Qiao}
 (Fellow, IEEE) received the B.Eng. degree in hydraulics and control and the M.Eng. degree in robotics and automation from Xi’an Jiaotong University, Xi’an, China, in 1986 and 1989, respectively, and the Ph.D. degree in robotics control from De Montfort University, Leicester, U.K., in 1995. 
Prof. Qiao is currently a Professor with the the State Key Laboratory of Multimodal Artificial Intelligence Systems, Institute of Automation, Chinese Academy of Sciences, Beijing, China. Her current research interests include robotics, machine learning, and pattern recognition. She is currently the member of the Chinese Academy of Science, and the Administrative Committee of the IEEE Robotics and Automation Society. She is the Editor in Chief of Assembly Automation, an Associate Editor of the IEEE Transactions on Cybernetics, IEEE Transactions on Neural Networks and Learning Systems, IEEE Transactions on Automation and Sciences Technology, IEEE Transactions on Cognitive and Developmental Systems, and IEEE/ASME Transactions on Mechatronics. 
\end{IEEEbiography}
 
\end{document}